\def\year{2022}\relax
\documentclass[letterpaper]{article} % DO NOT CHANGE THIS
\usepackage{aaai22}  % DO NOT CHANGE THIS
\usepackage{times}  % DO NOT CHANGE THIS
\usepackage{helvet}  % DO NOT CHANGE THIS
\usepackage{courier}  % DO NOT CHANGE THIS
\usepackage[hyphens]{url}  % DO NOT CHANGE THIS
\usepackage{graphicx} % DO NOT CHANGE THIS
\usepackage{natbib}  % DO NOT CHANGE THIS AND DO NOT ADD ANY OPTIONS TO IT
\usepackage{caption} % DO NOT CHANGE THIS AND DO NOT ADD ANY OPTIONS TO IT
\DeclareCaptionStyle{ruled}{labelfont=normalfont,labelsep=colon,strut=off} % DO NOT CHANGE THIS
\usepackage{algorithm}
\usepackage{algorithmic}
\usepackage{amsmath}

\usepackage{booktabs}

\usepackage{subfigure}

\usepackage{newfloat}
\usepackage{listings}
\floatstyle{ruled}
\newfloat{listing}{tb}{lst}{}
\floatname{listing}{Listing}

\title{FusionDepth: Complement Self-Supervised Monocular Depth Estimation with Cost Volume}
\author{
	Zhuofei Huang\textsuperscript{\rm 1}, Jianlin Liu\textsuperscript{\rm 2}, Shang Xu\textsuperscript{\rm 2}, Ying Chen\textsuperscript{\rm 2}, Yong Liu\textsuperscript{\rm 2}
}
\affiliations{
	\textsuperscript{\rm 1}HKUST\\
	\textsuperscript{\rm 2}Tencent\\
	zhuangbr@cse.ust.hk, \{jenningsliu, shangxu, mumuychen, choasliu\}@tencent.com
}

\usepackage{bibentry}
\begin{document}

\maketitle

\begin{abstract}

Multi-view stereo depth estimation based on cost volume usually works better than self-supervised monocular depth estimation except for moving objects and low-textured surfaces. So in this paper, we propose a multi-frame depth estimation framework which monocular depth can be refined continuously by multi-frame sequential constraints, leveraging a Bayesian fusion layer within several iterations. Both monocular and multi-view networks can be trained with no depth supervision. Our method also enhances the interpretability when combining monocular estimation with multi-view cost volume. Detailed experiments show that our method surpasses state-of-the-art unsupervised methods utilizing single or multiple frames at test time on KITTI benchmark. 
\end{abstract}

\section{Introduction}
% 1. 深度估计的重要性，强调depth评价的维度，不仅要准确，而且要完整
Predicting scene depth from imagery is of central importance for applications ranging from autonomous driving, robot navigation, to augmented reality. As acquiring depth ground truth for each pixel is challenging, self-supervised learning method with relative pose estimation as supervised signal is promising in monocular depth estimation. Meanwhile, with pose provided, multi-view cost volume aggregation is also a practical method to obtain accurate depth. A good depth estimation method should be capable of generating both accurate and complete depth.

\begin{figure} 
\centering
\includegraphics[width=1\columnwidth]{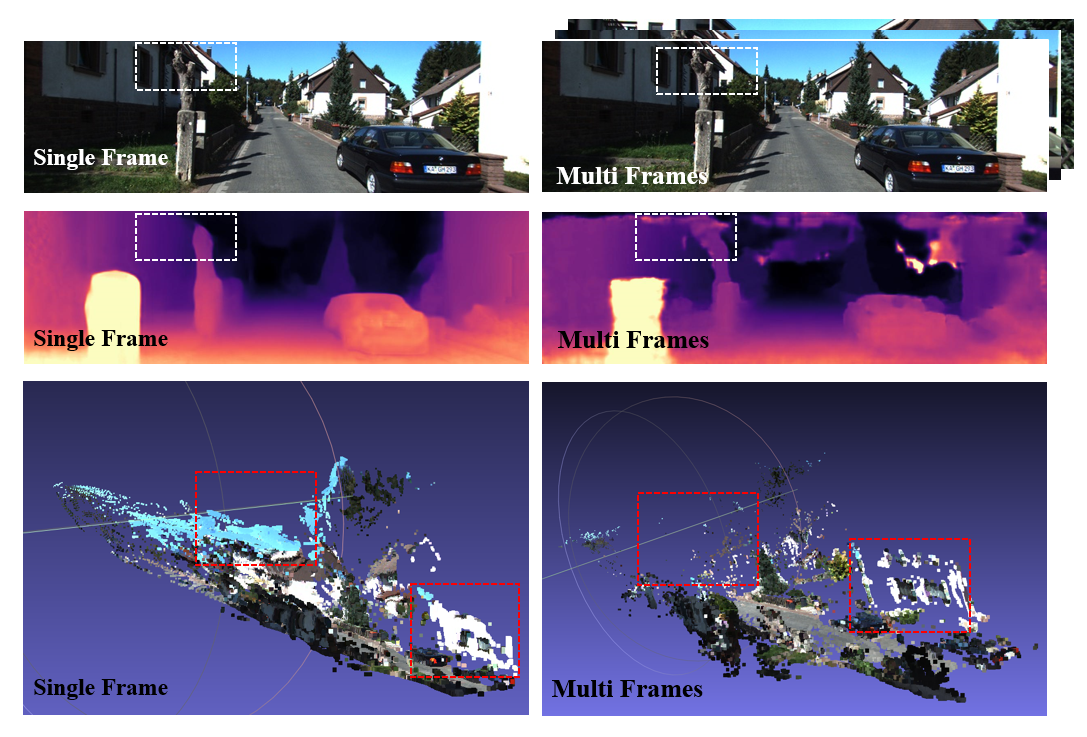}
\caption{Depth predictions from single (left column) and multi frames (right column). Multi-view method can predict high quality depth in highly textured regions as in white dotted boxes, while failing in low textured cases as in red boxes. In the third line we generate 3D maps from monocular and multi-frame methods which run on the whole video sequence. We find that less depth can pass depth consistency check for multi-frame methods in those untextured regions.}
\label{fig:figure1}
\end{figure}

% 2. 抛出问题：单目深度估计不准确，解释性不强；cost volome方法在弱纹理区域表现不好。希望结合。
However, monocular depth is not intepretable and hard to generalize. Due to the large domain gap between training and testing data, many pretrained networks in monocular depth estimation \cite{Zhou2017, Godard2019} may easily collapse in new scenes. Besides, monocular methods predict inaccurate depths for regions of large image gradient as shown in Figure \ref{fig:figure1}. 
Actually, in many practical scenarios such as camera on moving vehicle or handheld mobile phone, more than one frame is available at test time. Such multi-frame information is not exploited by recent monocular methods. To make use of additional frames at both training and testing time, MVSNet \cite{Yao2018} takes multiple images as input, and builds a cost volume with known camera poses, inferring depth in an end-to-end fashion. However, cost volume based methods degrade in moving object and untextured surfaces, resulting in incomplete depth maps as shown in Figure \ref{fig:figure1}. 
Intuitively, combining both monocular and multi-frame methods should generate both accurate and complete depth. 

% 3. 别人的解决方案 (不是related work，主要强调解决问题思路)，manydepth如何结合，monodepth和cost volume没有解耦。depth refinement的问题？我们怎么结合mono和cost volume，和上面的方案区别是什么，为什么会效果更好（解释性更强）
One recent approach Manydepth \cite{Watson2021} shows an implicit combination of monocular framework and cost volume. Different from MVSNet-like frameworks, it builds a cost volume concatenated with monocular feature map to regress final depths. It compares the depth represented by the $argmin$ of the cost volume with monocular depth to identify reliable pixels. However, this assumption is questionable for cost volume. In fact, \cite{Yang2020} shows explicit representation of depth output: expectation of probability volume. Besides, the distribution of probability volume reflects the confidence of multi-view stereo matching, which serves as an important criteria to find reliable pixels. Therefore, in this work we decouple cost volume from monocular network, and show how to exploit more information from cost volume and provide a reliable sparse depth map for following fusion step, with monocular depth. We called our multi-frame monocular fusion system \textbf{FusionDepth}. 

% 4. 总结我们的contribution
Our main contributions are listed below: 

\begin{itemize}
    \item We purpose a depth consistency check method to obtain reliable sparse depths from multi-view estimation. 
    \item Monocular depth with uncertainty will be continuously refined by sparse depths from different set of views by our purposed Bayesian fusion module.
    \item We introduce a strategy to further retrain monocular DepthNet with provided refined depths as supervision signal  during online adaptation learning. 
\end{itemize}

\section{Related Work}

% 2.1 单目方法综述
\subsection{Monocular Depth Estimation}

The main purpose of monocular depth estimation is to predict the depth of each pixel in a single input image. Supervised framework exploit dense supervision from depth sensors during training, e.g. \cite{Fu2018, Ramamonjisoa2020}. Since supervised methods perform poor generalization and have difficulty of accessing ground truth depth value, researchers tried self-supervised monocular depth estimation in recent years. 

Self-supervised methods usually train with image-reconstruction losses. \cite{Zhou2017} firstly purposes to adopt PoseCNN to estimate relative pose between each adjacent two frames. 
This work jointly updates depth and pose by minimizing image reconstruction loss. \cite{Bian2019} presents scale-consistent depth estimation with geometry consistency loss and a self-discovered mask for handling dynamic scenes. \cite{Godard2019} introduces auto-masking and min re-projection loss to solve the problems of moving objects and occlusion. By training with stereo pairs, \cite{Godard2017} introduces a left-right depth consistency loss. \cite{Vaishakh2020} incorporates Mirrored Exponential Disparity probability volumes to regress the expectation of depths and Mirror Occlusion Module to solve occlusion cases. 
Besides, for robustness to illumination change and occlusions, \cite{Klodt2018, Yang2020, Poggi2020} predict extra uncertainty map in self-supervised manner. 
Our monocular architecture is based on \cite{Godard2019, Poggi2020}, and provides initial guess of depth and uncertainty for all pixels. In contrast to previous monocular methods that directly regress final depth,  monocular depth will be continuously refined by online learning during our training process. 

% 2.2 多目方法综述
\subsection{Multi-frame Depth Estimation}
In recent years, end-to-end frameworks that leverage temporal information from multiple views at test time grow fast to help improve the quality of predicted depths. One general version of multi-view methods is to adopt RNN or LSTM-based models. \cite{Vaishakh2020} uses the architecture of the ConvLSTM for real-time self-supervised monocular depth estimation and completion. \cite{Wang2019} utilizes multi-view image reprojection and forward-backward flow consistency losses to train RNN model. Another version widely used as baseline is cost volume in multi-view stereo matching with known camera poses. MVSNet \cite{Yao2018} starts to present an end-to-end deep learning architecture for depth map inference from multi-view images. \cite{Gu2020} purposes a 3-stage 3D cost volume to generate multi-scale depth maps in a coarse-to-fine manner, where the range of depth interval reduces remarkably stage by stage. \cite{Zhang2020} regresses uncertainty map from probability volume to solve erroneous cost aggregation from occluded pixels. \cite{Huang2021} trains the MVSNet in an unsupervised manner by combining pixel-wise and feature-wise loss function, which helps decrease mismatch errors in some untextured and texture repeat areas. These works utilize sequence information and show better performance of depth estimation for most static regions with textured surfaces than monocular frameworks. 

In our work, MVSNet serves as real depth sensor. Assuming that ground truth of camera poses are known, a pretrained MVSNet provides estimation of depth values with distribution around the ground truth of depths. For the same referenced view, when we select different source views as input for MVSNet, we will generate distinct observations for depth value of each pixel in referenced view. 

% 2.3 融合方法综述
\subsection{Multi-frame Monocular Depth Fusion}

Combining both privileges of monocular and multi-frame depth estimation is an attractive technique. In fact it is necessary to generate both accurate and complete depth in most 3D reconstruction tasks, where monocular brings complete and multi-frame brings accuracy. 
\cite{Watson2021} predicts superior depths from a single image, or from multiple images when they are available, by concatenating monocular feature with cost volume. The network will select inliers by finding the $argmin$ of the cost volume, and encourage prediction of outliers to be similar to monocular estimation. 
\cite{Chen2019, Zhan2021, Li2021} introduces self-supervised framework for jointly learning depth and optical flow with online refinement strategies. In this paper, we purpose a novel \textbf{FusionDepth} which combines outputs from monocular network and Cascade MVSNet in Bayesian fusion module. The refinement takes effect in both training and testing, while training process is online learning.

\section{Method}
\noindent In this section, we introduce our depth fusion framework in detail. The system overview is illustrated in Fig. 2. Specifically, the goal of our system is to estimate a pixel-aligned depth map $D_t$ with adjacent 3 frames $I_{t-1}, I_t, I_{t+1}$ from video sequence. First, we apply a single-image depth estimation as introduced in monodepth2 \cite{Godard2019}, with an extra 2D-Convolution layer to predict depth uncertainty map $\sigma_t$. Then we use a representative MVSNet which contains 3-stage cascade cost volume to generate multi-view depth maps. Finally, combining single-image's uncertainty map with consistency checks of depths map from MVSNet, we adopt a Bayesian fusion module to refine depth values obtained from monocular depth maps.

% System overview框架图
\begin{figure*} 
\centering
\includegraphics[width=2\columnwidth]{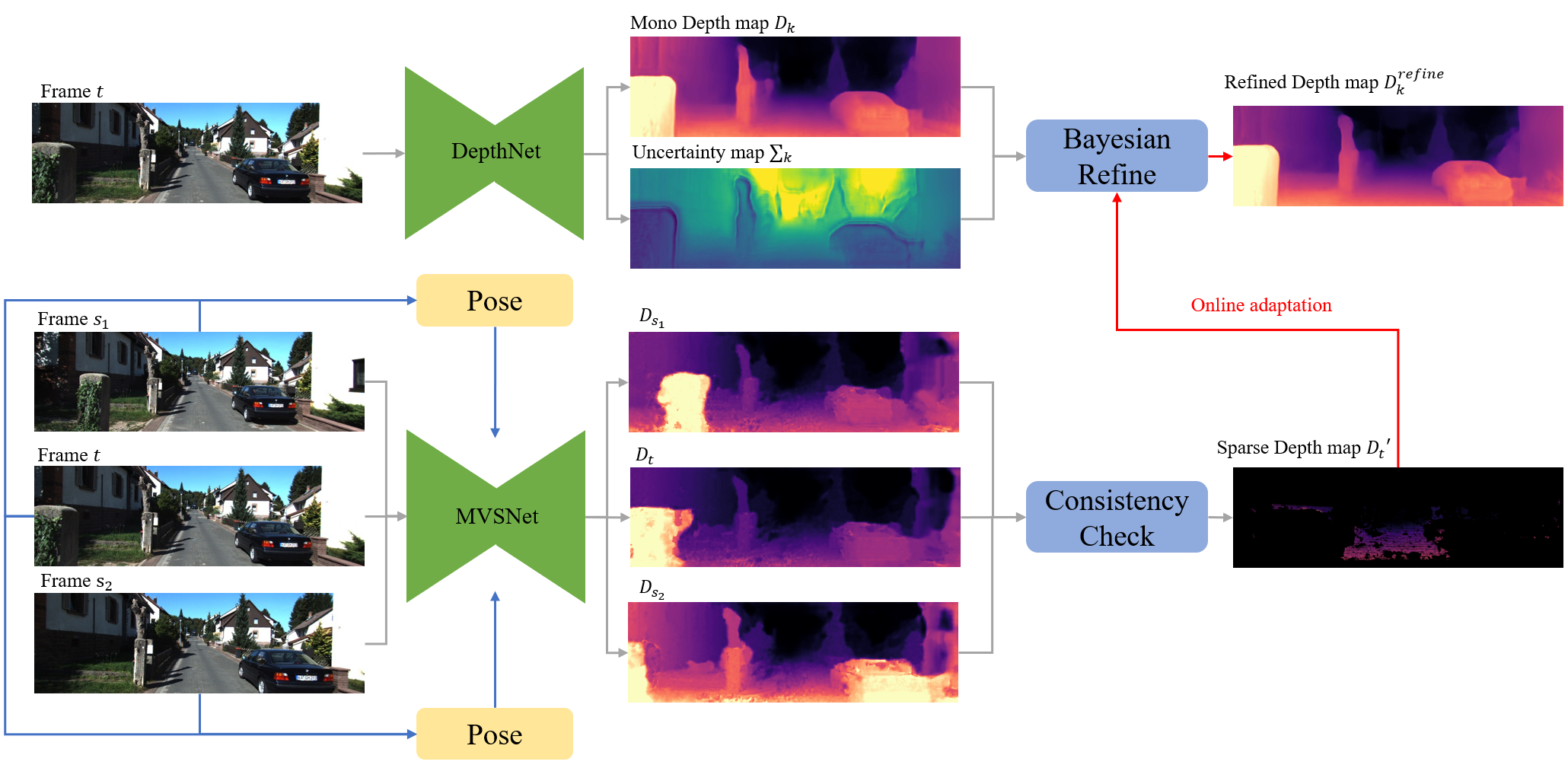}
\caption{The overview of our multi-view monocular depth fusion framework. Our final purpose is to estimate depth of frame $t$. First, DepthNet predicts single-view depth map $D_k$ and uncertainty map ${\Sigma_k}$ as initial guess. Then we select several source views $s_1, s_2, ...$, and feed them together with reference frame $t$ into MVSNet, which predicts depth map $D_{s_1}, D_t, D_{s_2}$ for each view. As we select different set of source views, depth result $D_t$ will differ. We take a multi-view depth consistency check on $D_{s_1}, D_t, D_{s_2}$ to retain inliers, and obtain the sparse projected depth $D_t'$ on frame $t$. Finally, $D_t'$ is used to refine initial monocular depth $D_k$ and uncertainty $\Sigma_k$ during an online Bayesian refinement module. So we obtain the final depth estimation $D_k^{refine}$ for frame $t$.}
\label{fig:eulerAngles}
\end{figure*}

\subsection{Monocular Depth Estimation}

\noindent Given one target image $I\in \mathcal{R}^{3\times H\times W}$, a monocular depth network $\mathcal F_D$ infers depth values for each pixel, i.e., $D=\mathcal F_D(I)$. $\mathcal F_D$ is designed as a skip-connection encoder-decoder architecture similar to Monodepth2 \cite{Godard2019}. Due to the lack of ground-truth depth values, we consider the target frame $I_t$ together with two adjacent source views $I_{t-1}, I_{t+1}$ to optimize $\mathcal F_D$ by warping pixels from source to target view and minimizing the photometric re-projection error:

\begin{equation}
\label{eq:pe}
r(I_t,I_{t'})=\frac{\alpha}{2}(1-SSIM(I_t,I_{t'}))+(1-\alpha)\Vert I_t-I_{t'} \Vert
\end{equation}

\noindent where $I_t'$ is the warped image from source view $I_s$ to target $I_t$, and $SSIM$ calculates structural similarity of patches. Similar to \cite{Godard2019} we follow the form of per-pixel minimum loss to tackle potential occlusion issues. The photometric loss is denoted as:

\begin{equation}
\label{eq:min_pe}
\mathcal L_{pho}=\min_t{r(I_t,I_{t'})}
\end{equation}

\noindent To enforce local smoothness, we adopt edge-aware loss for depth:

\begin{equation}
\label{eq:smooth}
\mathcal L_{smooth}=|\delta_xD_t|e^{-\delta_xI_t}+|\delta_yD_t|e^{-\delta_yI_t}
\end{equation}

\noindent where $\delta_x$ and $\delta_y$ represents gradients of depth and RGB images. The final loss of unsupervised depth estimation is:

\begin{equation}
\label{eq:depth_loss}
\mathcal L_{mono}=\frac{1}{s}\sum_i^s(\mathcal L_{pho}+\lambda\mathcal L_{smooth})
\end{equation}

\noindent where $s$ is the number of scales and $\lambda$ the weight of smooth term. 

\noindent In our system, We firstly pretrain DepthNet by minimizing $L_{mono}$ in (\ref{eq:depth_loss}). Besides, we predict an extra uncertainty map $\sigma_t$ by adding a $3\times 3$ convolution layer at the output of DepthNet. Notice that during pretraining, uncertainty $\sigma_t$ is not involved in $L_{mono}$, and it will be trained by Bayesian online learning introduced in the later section. 

% for the purpose of Bayesian Fusion in the later section, 

% \subsubsection{Uncertainty} DepthNet infers the scene depth $D$ for each pixel of $I$. The output depth map can usually be regarded as the mean of distribution $p(\hat{d}|d, \sigma)$ with respect to $\hat{d}$ in monocular depth estimation. Based on the framework above, we apply the Log-Likelihood Maximization strategy during training to predict heteroscedastic aleatoric uncertainty, to obtain depth mean and variance simultaneously. The key idea is to predict a posterior probability distribution for each pixel parameterized with its mean as well as its variance $p(\hat{d}|d, \sigma)$ over ground-truth labels $\hat{d}$. Assuming the noise of predictive distribution is modelled as Laplacian, the negative log-likelihood to be minimized is 

% \begin{equation}
% \label{eq:4_5}
% -\log{p(d|\hat{d}, \sigma)}=\frac{|d-\hat{d}|}{\sigma}+\log{\sigma}+const
% \end{equation}

% \noindent Note that no ground-truth label for $\sigma$ is required during training. The predicted uncertainty allows the network to adapt the weighting of the residual dependent on the data input, which improves the robustness of the model to noisy data or erroneous labels.

% \noindent The network is trained by log-likelihood maximization (i.e., negative log-likelihood minimization)

%%%%%%%%%%%%%%%%%%%%%%%%%%%%%%%%%%%%%%%%%%%%%%%%%%%%%%%%%%%%%%%%%%%%%%%%%%%%%%%%%%%%%%%%%%%%%%%%

% MVSNet 框架描述 Depth Estimation from Multi-View Framework
\subsection{Multi-View Depth Estimation}

\noindent To exploit multiple input frames in depth estimation, we build a 3-stage cascade cost volume \cite{Gu2020} which generates depth map in a coarse-to-fine manner. Given multiple input frames $I_t, I_{t-1}, I_{t+1}$, the Cascade MVSNet extracts 2D features for each view. We then warp the feature maps from source to target view $I_t$ using each of determined discrete hypothesis depth planes $\{d_j\}_{j=1}^{N_d}$, with the known camera intrinsics and relative poses. All warped features from source views are aggregated together into a single 3D probability volume via the \textit{soft-argmax} operation. Finally we regress the depth map $D$ and uncertainty map $U$ from the probability volume $\{P_{ij}\}_{j=1}^{N_d}$ for each pixel $i$, which follows the below:

%%%%%%%%%%%%%%%%%%%%%% MVSNet depth and uncertainty regression
\begin{equation}
\label{eq:expectation}
\begin{aligned}
D_i^{mvs}&=\sum_{j=1}^{N_d}d_jP_{ij}\\
U_i^{mvs}&=f_u(\sum_{j=1}^{N_d}-P_{ij}\log{P_{ij}})
\end{aligned}
\end{equation}

\noindent where depth value $D_i^{mvs}$ is computed as the expectation of distribution and uncertainty $U_i^{mvs}$ \cite{Yang2020} as the cross entropy over the probability distribution followed by a 2D CNN function layer $f_u$. Notice that the probability distribution $\{P_{ij}\}_{j=1}^{N_d}$ reflects the confidence of depth value for each pixel, which helps measure the similarity between corresponding image patches and determine whether they are matched. 

\subsubsection{Learn with Uncertainty}
Following the strategy in \cite{Yang2020, Poggi2020}, we jointly learning the depth $D_t^{mvs}$ and its uncertainty $U_t^{mvs}$ by minimizing the negative log likelihood of regular photometric loss:

\begin{equation}
\label{eq:nll_loss}
\mathcal L_{multi}=\frac{1}{V}\sum_{p_i\in V}\frac{\mathcal L_{pho}(I_i, D_i^{mvs})}{U_i^{mvs}}+\log{U_i^{mvs}}
\end{equation}

\noindent where uncertainty $U_i^{mvs}$ allows our network to adapt illumination change and occlusions \cite{Yang2020}. In most cases, MVSNet exploits sequence information so that we will get more accurate estimation of depth values than DepthNet, except the area of moving objects or low-textured surfaces. In the following sections we show how to combine these two methods and aggregate into a final refined depth map. 

\subsubsection{Depth Consistency Check}
Considering the predicted depth of target view $D_t^{mvs}$ is not reliable for all pixels, we adopt a simple multi-view depth consistency check strategy to select confident depth values. Given target view $I_t$ and $N$ source views $\{I_{s_i}\}_{i=1}^N$, we first compute depth maps of each source view by selecting it as target, while all other frames as source, and then feed them into cost volume to regress the depth maps $D_{s_i}^{mvs}$. Taking each source view and the target view as a pair $\{I_{s_i}, I_t\}$, with estimated depths and known relative pose, we can check the geometric consistency by considering two criteria: reprojection error and relative depth error. We first project all pixels from $I_t$ to $I_{s_i}$, and then re-project them back to $I_t$ with sampled source view depths. Besides, we warp the source depth $D_{s_i}$ to the target view as $D_{s_i\rightarrow t}^{mvs}$. We define the distance between reprojected pixel and the original one as $E_{dist}$, and the difference between reprojected depth map $D_{s_i\rightarrow t}^{mvs}$ and target depth map $D_t^{mvs}$ as $E_{diff}$. We consider depth value to be more confident if pixels in the target view satisfying $E_{dist}<e_1$ and  $E_{diff}<e_2$, by setting thresholds $e_1$ and $e_2$ as 1 and 0.001 respectively. In the following section, we will show how to apply $E_{dist}$ and $E_{diff}$ as the variance of a new observation for different inputs of MVSNet.

%
%
% 贝叶斯融合
\subsection{Bayesian Fusion}

% System overview框架图
\begin{figure} 
\centering
\includegraphics[width=1\columnwidth]{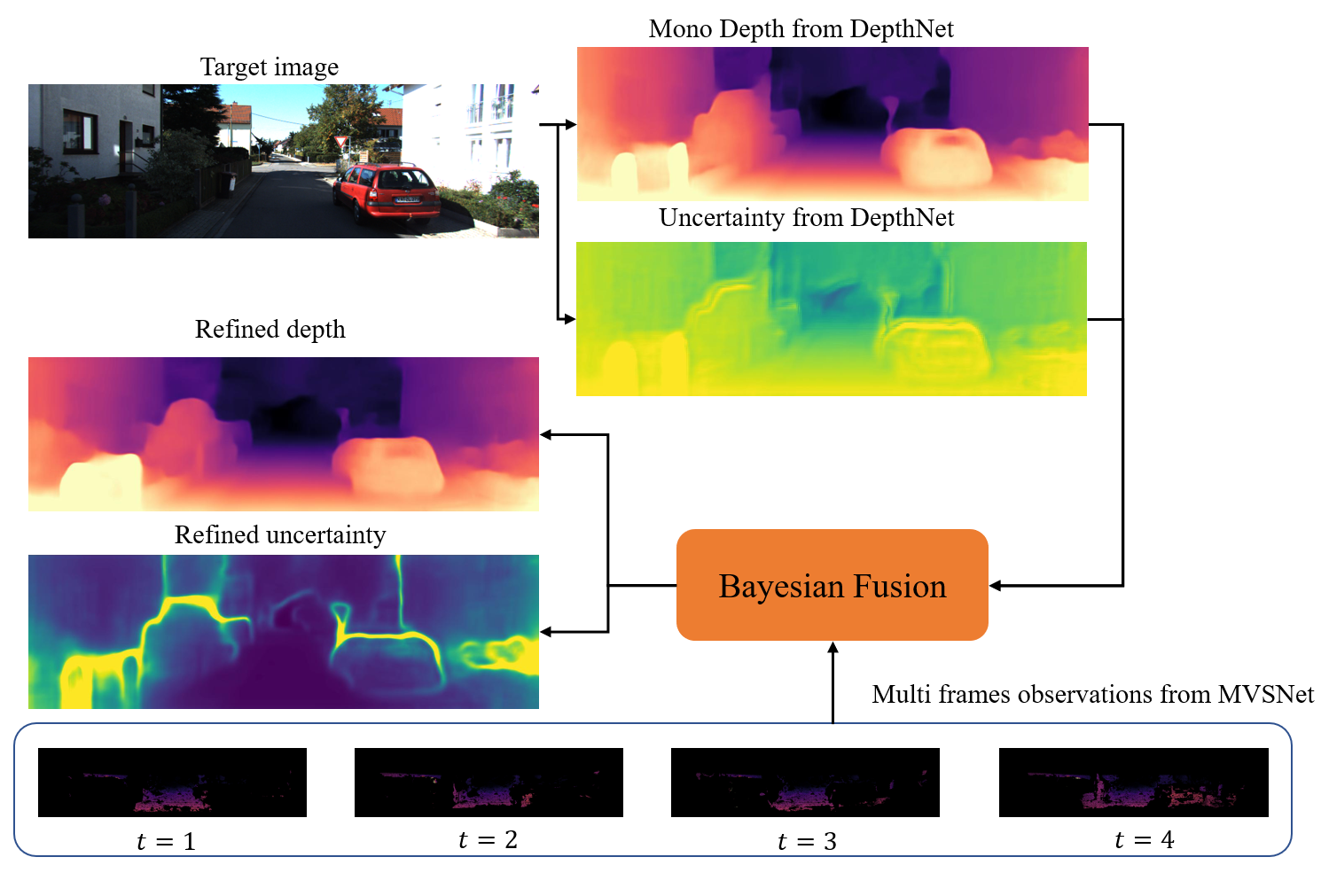}
\caption{Bayesian fusion process. When tested on KITTI, the pretrained DepthNet predicts an initial guess of depth and uncertainty. With incoming observations from MVSNet, both depth and uncertainty will be continuously refined in Bayesian fusion module, where depths become more accurate with decreasing uncertainty. }
\label{fig:bayesian}
\end{figure}

\noindent We purpose a Bayesian Fusion Module which refines the monocular depth map with filtered sparse depth maps from MVSNet. We model the depth estimation as inverse depth as in SVO \cite{Forster2014ICRA}, which is more robust for distant objects. The measurement of inverse depth $z_i=\frac{1}{d_i}$ is modeled with a mixture model of Gaussian and Uniform distribution \cite{Forster2014ICRA}, where a good measurement is normally distributed around the ground-truth $z_i$ and an outlier measurement arises from a uniform distribution in the interval $[z_i^{min}, z_i^{max}]$. For the same reference view $I_r$, if we choose different source views together with $I_r$ and pass them through MVSNet, we will regress distinct depth maps. Each depth inference can be treated as \textit{new observation} for depth values in reference view $I_r$. For every new observation $z_i^t$:

\begin{equation}
\label{eq:distribution}
p(z_i^t| z_i,\rho_i^t)=\rho_i^t \mathcal N(z_i^t| z_i,\tau_i^2)+(1-\rho_i^t) \mathcal U(z_i^t| z_i^{min},z_i^{max})
\end{equation}

\noindent where $\rho_i^t$ is the inlier probability which is modelled as Beta distribution and $\tau_i^2$ the variance of a good observation that can be computed geometrically as in our depth consistency check. 

\noindent The main idea of Bayesian fusion is to find the Maximum A Posteriori (MAP) estimation of $z_i^t$ for each observation, which can be approximated by the product of a Gaussian distribution for $z$ and a Beta distribution for inlier ratio $\rho$: 

\begin{equation}
\label{eq:map}
q(z,\rho|a_t, b_t,\mu_t, \sigma_t^2)=\textit{Beta}(\rho| a_t, b_t) \mathcal N(z_t| \mu_t,\sigma_t^2)
\end{equation}

\noindent As shown in Fig. \ref{fig:bayesian}, our Bayesian fusion module will update 4 parameters: $\mu_t, \sigma_t^2, a_t, b_t$, within $N$ iterations where $N$ is the number of new observations. The single-view depth estimation $d_k$ provides depth prior and inverse depth uncertainty $u_k$ from monocular \textit{DepthNet}. 

\noindent The mean and variance of inverse depth can be updated by:

\begin{equation}
\label{eq:update}
\begin{aligned}
&\mu_t=C_1'm+C_2'\mu_{t-1}\\
&\sigma_t^2=C_1'(s^2+m^2)+C_2'(\sigma_{t-1}^2+\mu_{t-1}^2)-\mu_{t-1}^2\\
\end{aligned}
\end{equation}

\noindent where $C_1',C_2',s,m$ are computed as in \cite{VOGIATZIS2011434}. As seen in (\ref{eq:update}), each iteration will refine the inverse depth values of inliers $\mu_t$ with decreasing uncertainty $\sigma_t^2$. The inverse depth $z_t$ will tend to be updated close to real inverse depth $z_i$ once the uncertainty value $\sigma_t^2$ is lower than a threshold. The parameters are initialized as follows:

\begin{equation}
\label{eq:init}
\begin{aligned}
&\mu_i^0=\frac{1}{d_k}, \qquad \sigma_i^0=u_k, \qquad z_i^{max}=\mu_i^0+\sigma_i^0 \\
&z_i^{min}=max(\mu_i^0-\sigma_i^0, 10^{-6})
\end{aligned}
\end{equation}

\begin{algorithm}[tb]
\caption{Bayesian Updating algorithm}
\label{alg:algorithm}
\textbf{Input}: Sparse inverse depths $\{z_i^t\}_{t=0}^{N-1}$ of $N$ observations\\
\textbf{Parameter}: $\mu_i, \sigma_i^2, a_i, b_i$\\
% \textbf{Output}: Your algorithm's output
\begin{algorithmic}[1] %[1] enables line numbers
\STATE Initialize parameters using (\ref{eq:init}).
\WHILE{${\sigma_i^t}^2$ not converges}
\STATE Choose a new observation $z_t$ and obtain its variance $\tau$ by adopting depth consistency check.
\FOR{every pixel $i$}
\IF {pixel $i$ is an inlier}
\STATE Update $\mu_i^t, {\sigma_i^t}^2$ using (\ref{eq:update})
\ENDIF
\ENDFOR
\ENDWHILE
% \STATE \textbf{return} solution
\end{algorithmic}
\end{algorithm}

\subsection{Loss Function}
After $N$ iterations in Bayesian fusion, we obtain the final refined depth map $D^r=\frac{1}{\mu_i^{N-1}}$. As our monocular DepthNet predicts depth $D_t$ and uncertainty $\sigma_t$, we adopt negative log-likelihood loss together with photometric and smooth loss in (\ref{eq:depth_loss}) as the total self-supervised loss:

\begin{equation}
\label{eq:final_loss}
\begin{aligned}
\mathcal L_{total}=\frac{|1/D^r-1/D_t|}{\sigma_t}+\log{\sigma_t}+\mathcal L_{mono}
\end{aligned}
\end{equation}

\noindent Our training process contains two stages: In the first stage, we minimize $\mathcal L_{mono}$ in (\ref{eq:depth_loss}) to pretrain monocular DepthNet and $\mathcal L_{multi}$ in (\ref{eq:nll_loss}) for MVSNet as in unsupervised learning of depth estimation. In the second stage, for every time step when we obtain a new observation from MVSNet, we online retrain DepthNet by minimizing (\ref{eq:final_loss}) to refine depth of inliers, while $\mathcal L_{mono}$ retains the photometric consistency and smoothness. 

%%%%%%%%%%%%%%%%%%%%%%%%%%%%%%%%%%%%%%%%%%%%%%%%%%%%%%%%%%%%%%%%%%%%%%%%%%%%%%%%%%%%%%%%%%%%%%%%

\section{Experiments}

\subsection{Implementation details}

\subsubsection{Network Architectures} For monocular architecture, we adopt \cite{Godard2019} and add a $3\times 3$ convolution layer at the output to generate depth uncertainty map. The backbone of our MVSNet is a cascade architecture as in \cite{Gu2020} and we add a shallow 2D CNN to convert  the entropy map to uncertainty map. Regarding numerical stability \cite{Kendall2017}, both uncertainty outputs are modelled as the log-variance in order to avoid zero values of the variance.

\subsubsection{Learning Settings} We use training-time color and flip augmentations on images being fed to the depth networks. We implement our models on PyTorch and train them on 4 Telsa V100 GPUs. The images are resized to $192\times 640$ for KITTI dataset. The monocular DepthNet and MVSNet are both pretrained first in a self-supervised manner for $10^5$ iterations by minimizing (\ref{eq:min_pe}) and (\ref{eq:nll_loss}). We use the Adam Optimizer with $\beta_1=0.9$, $\beta_2=0.999$. We train both models for 20 epochs with a batch size of 12. The initial learning rate is $10^{-4}$ and decreased to $10^{-5}$ after 15 epochs. We set the $SSIM$ weight as $\alpha = 0.85$ and smooth loss weight as $\lambda=10^{-3}$. In Bayesian Fusion, we choose 4 source views for each target $I_t$, includes $I_{t-1}$, $I_{t-2}$, $I_{t+1}$, $I_{t+2}$. For each measurement, we select 2 of them as source views, together with target $I_t$ as inputs of MVSNet. The threshold of Depth Filter $e_1$ and $e_2$ are set as $1$ and $0.001$ respectively. The fusion step takes 4 iterations, with source views of $\{I_{t-1}, I_{t+1}\}, \{I_{t-2}, I_{t+1}\}, \{I_{t-1}, I_{t+2}\}, \{I_{t-2}, I_{t+2}\}$ for each iteration respectively.

\begin{table*}[t]
\centering

\begin{tabular}{cccccccc}
\hline
 & Abs Rel & Sq Rel  & RMSE & RMSE Log  & $\delta < 1.25$ & $\delta < 1.25^2$ & $\delta < 1.25^3$ \\
\hline
Sfm-learner & 0.327 & 3.113 & 9.522 & 0.403 & 0.423 & 0.701 & 0.848 \\
SC & 0.163 & 0.964 & 4.913 & 0.224  & 0.776 & 0.932 & 0.977 \\
Monodepth2 & 0.120 & 0.779 & 4.611 & 0.178 & 0.846 & 0.962 & 0.988 \\
\hline
\textbf{Ours (p.d.)} & 0.118 & 0.627 & 4.111 & 0.171 & 0.850 & 0.967 & 0.991 \\
\textbf{Ours (p.k.)} & \textbf{0.100} & \textbf{0.473} & \textbf{3.520} & \textbf{0.149} & \textbf{0.885} & \textbf{0.976} & \textbf{0.994} \\
% Manydepth & 0.120 & 0.779 & 4.611 & 0.178 & 0.846 & 0.962 & 0.988 \\
\hline
\end{tabular}
%}
\caption{Quantitative results of depth estimation on KITTI odometry set 09, 10 for distance up to 80m. For error evaluating indexes, Abs Rel, Sq Rel, RMSE and RMSElog, lower is better, and for accuracy evaluating indexes, $\delta < 1.25$, $\delta < 1.25^2$, $\delta < 1.25^3$, higher is better. Sfm-learner: \cite{Zhou2017}, SC: \cite{Bian2019}, Monodepth2: \cite{Godard2019}. \textbf{p.d.}: MVSNet pretrained on DTU. \textbf{p.k.}: MVSNet pretrained on KITTI. }
\label{table1}
\end{table*}

% 深度图结果
\begin{figure*} 
\centering
\includegraphics[width=2\columnwidth]{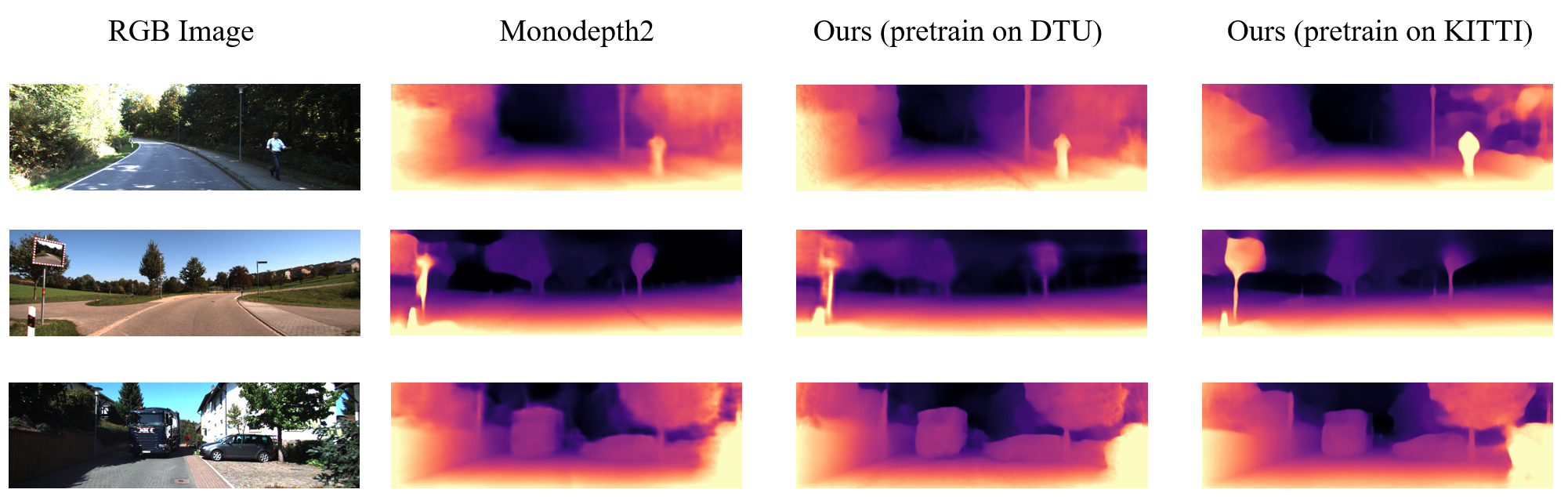}
\caption{\textbf{Qualitative monocular depth estimation performance} comparing our methods with previous monocular SOTA. We pretrain our network on DTU and KITTI, and test on KITTI. The results show that when adopting Bayesian fusion of our method to combine multi-view information, it will predict higher quality depths especially at edges as we apply Bayesian refinement even though the MVSNet model is pretrained in an unseen scene. }
\label{fig:Qualitative}
\end{figure*}

% uncertainty sparse curve
\begin{figure} 
\centering
\includegraphics[width=1\columnwidth]{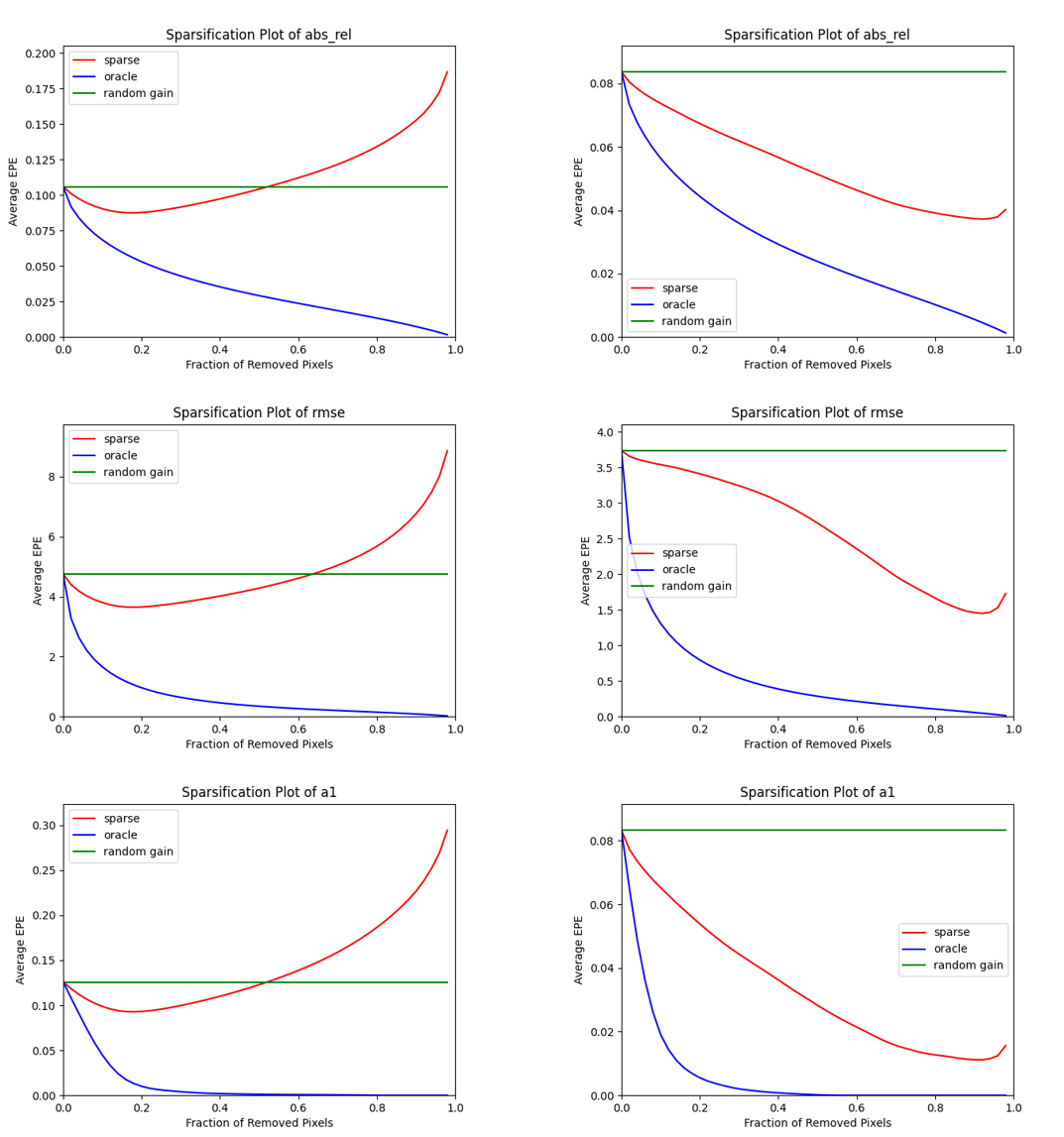}
\caption{\textbf{Sparsification curves} of predicted uncertainty. Left column: before refinement. Right column: after refinement.}
\label{fig:uncert_metrics}
\end{figure}

\subsection{Depth Estimation Performance} Since MVSNet strongly requires accurate camera poses to build cost volume, instead of testing on KITTI Eigen Split \cite{Eigen2014}, we adopt the KITTI odometry dataset which contains 11 driving sequences with ground-truth poses and depth available, except for 03 where ground-truth depth is not acquirable. We train the model on sequence 00-08 and evaluate the monocular depth estimation performance on 09 and 10. For 00-08, 35186 images are used for training and 3912 for validation. For 09-10, 2778 images are used for test. We compare our method with \cite{Zhou2017}, \cite{Bian2019}, \cite{Godard2019} on the same learning setting and dataset, and show the results in Table \ref{table1}. All groundtruth depth maps are capped at 80 meters. The result shows that our method with pretrained MVSNet on DTU and KITTI both outperforms SOTA monocular methods.

%%%%%%%%%%%%%%%%%%%%%%%%%%%%%%%%
\begin{figure}[t] 
\centering
    \subfigure{Pretrained on DTU: 153276 inliers}{
            \begin{minipage}[b]{0.45\textwidth}
                    \includegraphics[width=1\textwidth]{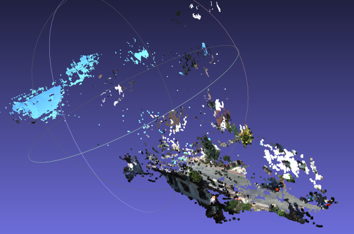}
            \end{minipage}
            \label{fig:pdtu}
        } 
        \subfigure{Pretrained on KITTI: 276464 inliers}{
                \begin{minipage}[b]{0.45\textwidth}
                   \includegraphics[width=1\textwidth]{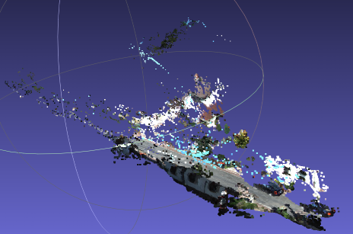}
                \end{minipage}
        \label{fig:pkitti}
        }
\caption{MVSNet pretrained on DTU and KITTI. A pretrained MVSNet model in unseen environment can generate sufficient inliers for Bayesian fusion even if less than the one pretrained in KITTI, after multi-view depth consistency check. }% with overlap predicted by our OETR.}}
\label{fig:3dpc}
\end{figure}
%%%%%%%%%%%%%%%%%%%%%%%%%%%%%%%%

\subsection{Ablation Studies}
To further explore the performance improvements that our network provides, we take some extra ablative analysis on the different components. 

\subsubsection{Uncertainty Metrics} Since we have estimated uncertainty map in our DepthNet, it is necessary for us to evaluate how significant the modelled uncertainties are. Intuitively, ideal uncertainty estimation should properly reveal the real error of predicted depth maps. Following the sparsification plots as in \cite{Ilg2018}, given a depth error metric $\eta$ (Eg., Abs Rel, RMSE, $\delta\ge 1.25$), we sort depths of all pixels in descending order of uncertainty. Iteratively, we remove a subset of pixels with highest uncertainty, and compute depth metric $\eta$ on the remaining pixels to plot a curve (as shown in red curve in Fig \ref{fig:uncert_metrics}, which is called 
\textit{Sparsification Curve}). As we remove 2\% pixels each time, the mean of depth metric on remaining pixels will decrease sharply if our uncertainty estimation is good. Obviously, the ideal \textit{Sparsification Curve} is obtained by sorting pixels in descending order of the $\eta$ magnitude, which is called \textit{Oracle Curve} as shown in blue curve in Fig \ref{fig:uncert_metrics}. Therefore, the closer between \textit{Sparsification Curve} (red) and \textit{Oracle Curve} (blue), the better our uncertainty estimation is. 

When we plot the curve with original total mean metrics, as seen in green curves named \textit{Random Curve}, we plot all of the three curves for each depth metric: (Abs Rel, RMSE, $\delta\ge 1.25$), before and after Bayesian refinement, as seen in Fig \ref{fig:uncert_metrics}. In the left column, an initial guess of uncertainty before refinement is of random high value for each pixel, so uncertainty does not show the similar distribution to depth metrics and \textit{Sparsification Curve} (red) is far away from \textit{Oracle} (blue). In the right column, after Bayesian refinement, uncertainty of many inliers decrease to a proper threshold, thus \textit{Sparsification Curve} seems closer to \textit{Oracle} (blue), which proves a conclusion that Bayesian fusion module can make our predicted uncertainty map more intepretable due to the consistency between uncertainty and real depth error.

\subsubsection{Generalization}
Distinct from monocular depth estimation learning with photometric error \cite{Godard2019}, learning-based multi-view stereo matching methods encourage network to match high-level features and show stronger robustness when inferring depth maps in an unseen scene. Thus, it is possible for us to pretrain MVSNet on another unseen indoor scene with ground truth depth provided, and then test on KITTI while taking Bayesian fusion steps. Since during the whole training process, we do not use any ground truth depth of KITTI, our training is still self-supervised. By adopting the pretrained model from DTU \cite{Yao2018} where MVSNet converges smoothly, we take multi-view depth consistency check on the output depth maps when testing on KITTI, the result is shown in Fig \ref{fig:3dpc}. The model pretrained on DTU provides roughly half inliers of that pretrained on KITTI, the former number of inliers is still large for refinement. As seen in Table \ref{table1} again, our method with pretrained model on DTU still outperforms the monocular ones. 

% \begin{tabular}{cccccccc}
% \hline
%  & Abs rel & Sq Rel  & RMSE & RMSE Log & $\delta < 1.25$ & $\delta < 1.25^2$ & $\delta < 1.25^3$ \\
% \hline
% Monodepth2 & 0.120 & 0.779 & 4.611 & 0.178 & 0.846 & 0.962 & 0.988 \\
% \hline
% \end{tabular}

%%%%%%%%%%%%%%%%%%%%%%%%%% Conclusions %%%%%%%%%%%%%%%%%%%%%%%%%%
\section{Conclusions}
We propose an online learning multi-frame monocular combined system for depth estimation with the help of multi-view stereo matching and Bayesian fusion. The predicted single-view depth is continuously refined with incoming output from cost volume with different source views as input. The updating process follows Bayesian formula and our final refined depth will converge to the ground truth once the refined uncertainty decreases to a threshold. Our system leverages the generalization ability of cost volume to make up for the shortage of monocular depth estimation, so that we achieve the purpose of generating both accurate and complete depth.

%%%%%%%%%%%%%%%%%%%%%%%%%% End %%%%%%%%%%%%%%%%%%%%%%%%%%
\bibliography{ref.bib}

\begin{thebibliography}{27}
\providecommand{\natexlab}[1]{#1}

\bibitem[{Bian et~al.(2019)Bian, Li, Wang, Zhan, Shen, Cheng, and
  Reid}]{Bian2019}
Bian, J.~W.; Li, Z.; Wang, N.; Zhan, H.; Shen, C.; Cheng, M.~M.; and Reid, I.
  2019.
\newblock {Unsupervised scale-consistent depth and ego-motion learning from
  monocular video}.
\newblock In \emph{NeurIPS}.

\bibitem[{Chen, Schmid, and Sminchisescu(2019)}]{Chen2019}
Chen, Y.; Schmid, C.; and Sminchisescu, C. 2019.
\newblock {Self-supervised learning with geometric constraints in monocular
  video: Connecting flow, depth, and camera}.
\newblock In \emph{ICCV}.

\bibitem[{David~Eigen and Fergus(2014)}]{Eigen2014}
David~Eigen, C.~P.; and Fergus, R. 2014.
\newblock {Depth Map Prediction from a Single Image using a Multi-Scale Deep
  Network}.
\newblock arXiv:1406.2283.

\bibitem[{Forster, Pizzoli, and Scaramuzza(2014)}]{Forster2014ICRA}
Forster, C.; Pizzoli, M.; and Scaramuzza, D. 2014.
\newblock {SVO}: Fast Semi-Direct Monocular Visual Odometry.
\newblock In \emph{ICRA}.

\bibitem[{Fu et~al.(2018)Fu, Gong, Wang, Batmanghelich, and Tao}]{Fu2018}
Fu, H.; Gong, M.; Wang, C.; Batmanghelich, K.; and Tao, D. 2018.
\newblock {Deep Ordinal Regression Network for Monocular Depth Estimation}.
\newblock In \emph{CVPR}.

\bibitem[{Godard, Aodha, and Brostow(2017)}]{Godard2017}
Godard, C.; Aodha, O.~M.; and Brostow, G.~J. 2017.
\newblock {Unsupervised Monocular Depth Estimation with Left-Right
  Consistency}.
\newblock In \emph{CVPR}.

\bibitem[{Godard et~al.(2019)Godard, Aodha, Firman, and Brostow}]{Godard2019}
Godard, C.; Aodha, O.~M.; Firman, M.; and Brostow, G. 2019.
\newblock Digging into self-supervised monocular depth estimation.
\newblock In \emph{ICCV}.

\bibitem[{Gu et~al.(2020)Gu, Fan, Zhu, Dai, Tan, and Tan}]{Gu2020}
Gu, X.; Fan, Z.; Zhu, S.; Dai, Z.; Tan, F.; and Tan, P. 2020.
\newblock Cascade Cost Volume for High-Resolution Multi-View Stereo and Stereo
  Matching.
\newblock In \emph{CVPR}.

\bibitem[{Huang et~al.(2021)Huang, Yi, Huang, He, Liu, and Liu}]{Huang2021}
Huang, B.; Yi, H.; Huang, C.; He, Y.; Liu, J.; and Liu, X. 2021.
\newblock {M3VSNet: Unsupervised multi-metric multi-view stereo network}.
\newblock In \emph{2021 IEEE International Conference on Image Processing
  (ICIP)}, 3163--3167. IEEE.

\bibitem[{Ilg et~al.(2018)Ilg, {\c{C}}i{\c{c}}ek, Galesso, Klein, Makansi,
  Hutter, and Brox}]{Ilg2018}
Ilg, E.; {\c{C}}i{\c{c}}ek, {\"{O}}.; Galesso, S.; Klein, A.; Makansi, O.;
  Hutter, F.; and Brox, T. 2018.
\newblock {Uncertainty estimates and multi-hypotheses networks for optical
  flow}.
\newblock In \emph{ECCV}.

\bibitem[{Kendall and Gal(2017)}]{Kendall2017}
Kendall, A.; and Gal, Y. 2017.
\newblock {What uncertainties do we need in Bayesian deep learning for computer
  vision?}
\newblock In \emph{NeurIPS}.

\bibitem[{Klodt and Vedaldi(2018)}]{Klodt2018}
Klodt, M.; and Vedaldi, A. 2018.
\newblock {Supervising the new with the old: Learning SFM from SFM}.
\newblock In \emph{ECCV}.

\bibitem[{Li et~al.(2021)Li, Wu, Cao, and Zha}]{Li2021}
Li, S.; Wu, X.; Cao, Y.; and Zha, H. 2021.
\newblock {Generalizing to the Open World: Deep Visual Odometry with Online
  Adaptation}.
\newblock In \emph{CVPR}.

\bibitem[{Patil et~al.(2020)Patil, Gansbeke, Dai, and Gool}]{Vaishakh2020}
Patil, V.; Gansbeke, W.~V.; Dai, D.; and Gool, L.~V. 2020.
\newblock {Don't Forget The Past: Recurrent Depth Estimation from Monocular
  Video}.
\newblock arXiv:2001.02613.

\bibitem[{Poggi et~al.(2020)Poggi, Aleotti, Tosi, and Mattoccia}]{Poggi2020}
Poggi, M.; Aleotti, F.; Tosi, F.; and Mattoccia, S. 2020.
\newblock On the uncertainty of self-supervised monocular depth estimation.
\newblock In \emph{CVPR}.

\bibitem[{Ramamonjisoa, Du, and Lepetit(2020)}]{Ramamonjisoa2020}
Ramamonjisoa, M.; Du, Y.; and Lepetit, V. 2020.
\newblock {Predicting sharp and accurate occlusion boundaries in monocular
  depth estimation using displacement fields}.
\newblock In \emph{CVPR}.

\bibitem[{Srivastava et~al.(2014)Srivastava, Hinton, Krizhevsky, Sutskever, and
  Salakhutdinov}]{srivastava14a}
Srivastava, N.; Hinton, G.; Krizhevsky, A.; Sutskever, I.; and Salakhutdinov,
  R. 2014.
\newblock {Dropout: A Simple Way to Prevent Neural Networks from Overfitting}.
\newblock \emph{Journal of Machine Learning Research}, 15(56): 1929--1958.

\bibitem[{Uhrig et~al.(2017)Uhrig, Schneider, Schneider, Franke, Brox, and
  Geiger}]{Uhrig2017}
Uhrig, J.; Schneider, N.; Schneider, L.; Franke, U.; Brox, T.; and Geiger, A.
  2017.
\newblock {Sparsity Invariant CNNs}.
\newblock arXiv:1708.06500.

\bibitem[{Vogiatzis and Hernández(2011)}]{VOGIATZIS2011434}
Vogiatzis, G.; and Hernández, C. 2011.
\newblock Video-based, real-time multi-view stereo.
\newblock \emph{Image and Vision Computing}, 29(7): 434--441.

\bibitem[{Wang, Pizer, and Frahm(2019)}]{Wang2019}
Wang, R.; Pizer, S.~M.; and Frahm, J.-M. 2019.
\newblock {Recurrent Neural Network for (Un-)supervised Learning of Monocular
  VideoVisual Odometry and Depth}.
\newblock arXiv:1904.07087.

\bibitem[{Watson et~al.(2021)Watson, {Mac Aodha}, Prisacariu, Brostow, and
  Firman}]{Watson2021}
Watson, J.; {Mac Aodha}, O.; Prisacariu, V.; Brostow, G.; and Firman, M. 2021.
\newblock {The Temporal Opportunist: Self-Supervised Multi-Frame Monocular
  Depth}.
\newblock In \emph{CVPR}.

\bibitem[{Yang et~al.(2020)Yang, {Von Stumberg}, Wang, and Cremers}]{Yang2020}
Yang, N.; {Von Stumberg}, L.; Wang, R.; and Cremers, D. 2020.
\newblock {D3VO}: Deep Depth, Deep Pose and Deep Uncertainty for Monocular
  Visual Odometry.
\newblock In \emph{CVPR}.

\bibitem[{Yao et~al.(2018)Yao, Luo, Li, Fang, and Quan}]{Yao2018}
Yao, Y.; Luo, Z.; Li, S.; Fang, T.; and Quan, L. 2018.
\newblock {MVSNet: Depth inference for unstructured multi-view stereo}.
\newblock In \emph{ECCV}.

\bibitem[{Zhan et~al.(2020)Zhan, Weerasekera, Bian, Garg, and Reid}]{Zhan2021}
Zhan, H.; Weerasekera, C.~S.; Bian, J.-W.; Garg, R.; and Reid, I. 2020.
\newblock {DF-VO: What Should Be Learnt for Visual Odometry?}
\newblock In \emph{ICRA}.

\bibitem[{Zhang et~al.(2020)Zhang, Yao, Li, Luo, and Fang}]{Zhang2020}
Zhang, J.; Yao, Y.; Li, S.; Luo, Z.; and Fang, T. 2020.
\newblock Visibility-aware Multi-view Stereo Network.
\newblock arXiv:2008.07928.

\bibitem[{Zhou et~al.(2017)Zhou, Brown, Snavely, and Lowe}]{Zhou2017}
Zhou, T.; Brown, M.; Snavely, N.; and Lowe, D. 2017.
\newblock Unsupervised Learning of Depth and Ego-Motion from Video.
\newblock In \emph{CVPR}.

\bibitem[{Zou, Luo, and Huang(2018)}]{Zou2018}
Zou, Y.; Luo, Z.; and Huang, J.-B. 2018.
\newblock {DF-Net: Unsupervised Joint Learning of Depth and Flow using
  Cross-Task Consistency}.
\newblock In \emph{ECCV}.

\end{thebibliography}
\end{document}